# RSwinV2-MD: An Enhanced Residual SwinV2 Transformer for Monkeypox Detection from Skin Images

Rashid Iqbal[1], Saddam Hussain Khan[1]*

[1]Artificial Intelligence Lab, Department of Computer Systems Engineering, University of Engineering and Applied Sciences (UEAS), Swat 19060, Pakistan

**Email:** hengrshkhan822@gmail.com

## ABSTRACT

Monkeypox (MPox) has recently become a rising global health threat due to its efficiency in being transmitted via direct contact with the skin. The accuracy of lesion observation in its early stages has been critical in containing the outbreak of the virus; however, due to its high infectivity and development of new strains, monitoring and screening have become more difficult. In this paper, a deep learning approach for Mpox diagnosis named "Customized Residual SwinTransformerV2 (RSwinV2) has been proposed, trying to enhance the capability of lesion classification by employing the RSwinV2 tool-assisted vision approach. In the RSwinV2 method, a hierarchical structure of the transformer has been customized based on the input dimensionality, embedding structure, and output targeted by the method. In this RSwinV2 approach, the input image has been split into non-overlapping patches and processed using shifted windows and attention in these patches. This process has helped the method link all the windows efficiently by avoiding the locality issues of non-overlapping regions in attention, while being computationally efficient. RSwinV2 has further developed based on SwinTransformer and has included patch and position embeddings to take advantage of the transformer's global-linking capability by employing multi-head attention in these embeddings. Furthermore, RSwinV2 has developed and incorporated the Inverse Residual Block (IRB) into this method, which utilizes convolutional skip connections with these inclusive designs to address the vanishing gradient issues during processing. RSwinV2's inclusion of IRB has therefore facilitated this method to link global patterns as well as local patterns; hence, its integrity has helped improve lesion classification capability by minimizing variability of Mpox and increasing differences of Mpox, chickenpox, measles, and cowpox. In testing RSwinV2, its accuracy of 96.51% and F1 of 96.13% have been achieved on the Kaggle public dataset, which has outperformed standard CNN models and SwinTransformers; RSwinV2's vector has thus proved its valiance as a computer-assisted tool for Mpox lesion observation interpretation.

1. Introduction:

Monkeypox (MPox), of Orthopoxvirus, has a strong zoonotic relationship with smallpox and vaccinia viruses; thus, MPox has shown potential re-emerging outbreaks and local epidemics [1], [2]. Since it was first recognized in 1959, MPox has existed in human populations since 1970, with the first identified case in a human host [3], [4]. Even though MPox is far less virulent than COVID-19 [5], a significant rise in reported cases has been shown worldwide [3], [6]. The mode of infection is direct contact with an already-infected human or animal, or indirect contact via exposure to contaminated surfaces. Fever, myalgias, and fatigue often follow in the wake of onset, but the characteristic rash-like lesions on the skin remain the feature of choice and focus of imaging assessment [7].

There are challenges in automatically diagnosing MPox from medical images. For one, high-dimensional representations of images, in addition to the limited amount of labeled training examples, might worsen the vulnerability to overfitting and the curse of dimensionality [8]. Transfer learning is often utilized to overcome these challenges in limited training examples [4], [9]. Furthermore, differences in lesion appearance, location, and contrast, as well as similarities with eruptive diseases, make it difficult to distinguish correctly. Traditional vision transformers might not work properly in capturing delicate details in local features and tend to consume high computational power [10], [11].

However, these disadvantages recommend the development of a Swin Transformer model developed for accurate MPox lesion detection. The model is based on a hierarchical transformer model in which the image is divided into nonoverlapping patches with a shifted window self-attention operation. Such a network would allow information to interact across windows with reduced computational cost and would not suffer from the locality constraint created due to the nonoverlapping window constraint introduced in traditional attention windows. The future scope will be to leverage this model to better detect MPox-related skin lesions. With this background, the major contributions of this work are:

- The proposed RSwinV2 is customized and adapted to the input data dimensionality, embedding structure, and outcome of interest targeted in MPox image analysis. The proposed RSwinV2 framework consists of the embedding of patches and positions, multi-head attention mechanisms to capture global dependencies, and uses an inverse residual block.
- The IRB uses skip connections with convolution layers to efficiently encode lesion characteristics through locally correlated features and avoid vanishing gradients.
- The demonstration that RSwinV2 facilitates joint learning of long-range and local patterns. This dual mechanism reduces intra-class variability in MPox presentations and enhances discrimination from visually similar diseases, including chickenpox, measles, and cowpox.

- Empirical validation showing that the IRB enhances local feature extraction while the shifted-window multi-head self-attention effectively captures global context, collectively contributing to superior diagnostic performance compared to benchmark models.
- The proposed RSwinV2 framework demonstrates superior performance, achieving the highest classification accuracy on both the Kaggle benchmark and diverse multi-source datasets when compared to state-of-the-art CNN and Vision Transformer models.

The manuscript is structured as follows: Section 2 provides a view of related works. Section 3 introduces the new MPox diagnostic system. Section 4 mentions the datasets, data pre-processing methods, and evaluation criteria. Section 5 displays the results of experiments. At last, Section 6 closes this paper with proposals for further research.

## 2. Literature Review

Deep learning (DL) was found to have efficacy in medical imaging and has been employed in diverse clinical settings, including brain cancers, pneumonia, tuberculosis, and COVID-19, as established in previous research works [12]–[14]. The rising incidence of Monkeypox (MPox), as well as limitations in testing and specialists in some regions, has thus driven the need for a decision support system in imaging [11].

The early works on MPox classification were dominantly conducted using convolutional neural network (CNNs) architectures, as represented in Table 1. MobileNetV2 and VGG derivatives were considered for MPox image classification [15]. AlexNet and VGG16/VGG19 were considered on digital skin images [16]. Custom-built pipelines using DenseNet-201 were also documented, including an evaluation using DenseNet-201, which reported results [17]. Lightweight networks with an attention mechanism and deeper residual learning architectures were also considered, including attention MobileNetV2, M-ResNet50, and DarkNet53 on MPox publicly available datasets [18]–[20].

Vision Transformers (ViTs) recently emerged as a solution to address the drawbacks of strictly local CNN-based feature learning in images [21]. On the MPox image datasets, the application of the ViT concept as well as the hybrid CNN-ViT models has led to competitive results, with the contribution of the ViT on the MSLD and other public datasets, as discussed in references [22]–[25]. Additionally, models incorporating a CNN feature extractor and the transformer aggregator, as in the case of the Bagging-Ensemble of the DenseNet201-ViT, have been discussed in reference [26].

Nevertheless, some of the ongoing challenges to MPox image classification, even in the face of reported advancements, have been mentioned in the literature as: the potential underestimation of inter-pixel dependencies by CNN-based architectures, the ViT model's potential susceptibility to a patch arrangement, or the lack of preservation of local details [27]–[29]. The computational expense, interpretability, as well as the ability to generalize well across datasets have been mentioned as ongoing issues in the literature, as well [30] and [27]. All these issues have been addressed by the devised hierarchical transformer.

Table 1. Earlier research on MPox detection has utilized CNNs, ViTs, and hybrid techniques.

| Author (Year) | Dataset Detail | Model | Acc |
|---|---|---|---|
| Ali et al; (22). [31] | MPox Skin Lesion Dataset (MSLD). | ResNet-50 | 82.96 |
| Irmak et al; (22). [15] | MSLD. | MobileNet-V2 | 91.38 |
| Sahin et al; (22). [32] | MSLD. | MobileNet-V2 | 91.11 |
| Sorayaie Azar et al; (23). [33] | MPox-dataset-2022 | DenseNet-201 | 95.18 |
| Bala et al; (23). [17] | MPox-Skin_ImagesDataset (MSID) | DenseNet-201 | 93.19 |
| Eliwa et al; (23). [34] | Mpox-PATIENTS | GWO-optimized CNN's | 95.30 |
| Biswas et al; (24). [20] | MSLD-V2 | DarkNet-53 | 85.78 |
| M. Ahsan et al; (24). [19] | Customized | Hybrid (CNN-ViT) | 89.00 |
| Deb Raha; (24). [18] | MSID. | Attention-based MobileNet-V2 | 92.28 |
| Oztel; (24). [26] | PAD_UFES_20 + MSLD | Ensemble (ViT + Densenet-201) | 81.91 |
| Arshed_et_al. (24). [25] |  | ViT's | 93.00 |
| Gizachew_Mulu_Setegn (25) [35] | Git-Hub | LGBM_Classifier | 89.30 |

## 3. Methodology

The proposed RSwinV2 model aims to improve its feature representation abilities for automatic MPox diagnosis. An overview of the pipeline is presented in Figure 1, which includes image preprocessing and data augmentation, followed by lesion classification using a customized Swin Transformer backbone. Performance comparison is performed against established CNN and transformer baselines.

### 3.1. Data Preprocessing and Augmentation

All images are rescaled to a fixed resolution and are standardized through intensity normalization to cancel out variability introduced by the image acquisition process. Data augmentation is also implemented during the training phase to generalize well in scenarios where the amount of data is very scarce, as is the case for many datasets involving medical images. Data augmentation is implemented through transformations that are spatial and photometric and are aimed at enhancing the semantic understanding of the lesions while introducing variability into the inputs to cancel out the overfitting problem.

### 3.2. Proposed RSwinV2 Architecture

RSwinV2 is a customised hierarchical Swin Transformer framework for MPox image analysis, designed to jointly capture global contextual relationships and fine-grained local lesion patterns [28]. Input images are standardised to a fixed tensor size and converted into a sequence of non-overlapping patch tokens, where a linear projection maps patch vectors into a learnable embedding space and the classification head maps the final representation to five outcome classes. Shifted-window self-attention enables cross-window information exchange with computational efficiency, consequently mitigating locality constraints associated with non-overlapping attention windows.

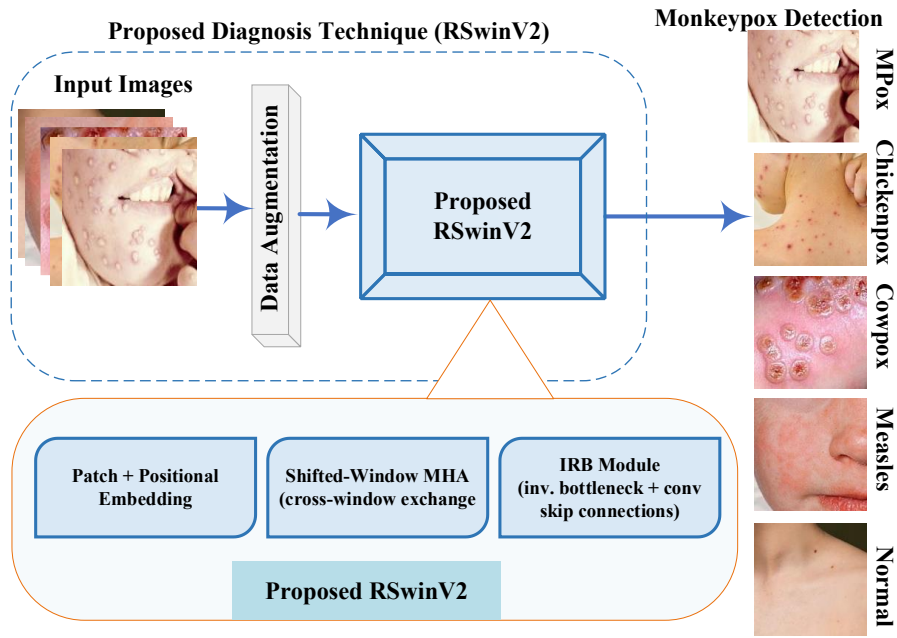

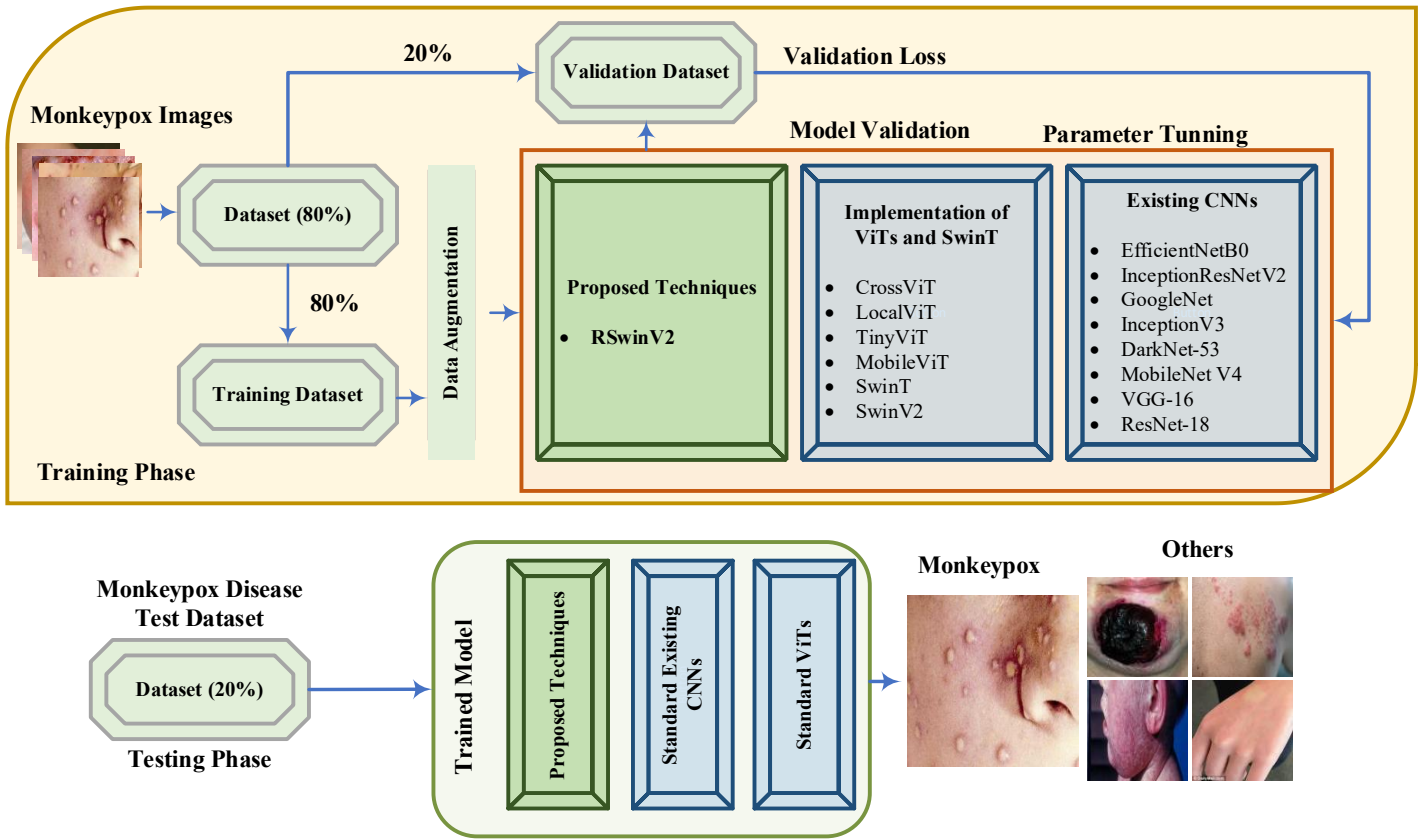

Figure 1: An overview of the pipeline.

Each transformer block comprises (i) a patch and positional embedding, (ii) multi-head self-attention (MHA) for global dependency modelling, and (iii) an inverse residual block (IRB) that replaces the standard feed-forward network, as illustrated in Figure 2. The IRB integrates convolutional skip connections within an inverted bottleneck design, hence strengthening local feature extraction and supporting stable gradient flow within the transformer stack.

### 3.2.1. Patch and Positional Embedding

Let an RGB input image be $I \in R^{H \times W \times 3}$. Patchification reshapes the image into (N) flattened patches of size (P x P) as shown in equation 1, where

$$N = \left(\frac{H}{P}\right)\left(\frac{W}{P}\right), D = P^2 \qquad (1)$$

In the experimental configuration, images are resized to 224 x 224 x 3 and partitioned into 16 x 16 patches, yielding (N=196) tokens. Patch vectors are linearly projected into a d-dimensional embedding space represented in equation 2 via $E \in R^{D \times d}$. A learnable positional embedding $E_{pos} \in R^{(N+1) \times d}$ is added to preserve spatial ordering, and a learnable classification token $E_{cls} \in R^d$ is prepended to the sequence for classification, is signified in equation 3 as below.

$$X_{patch} \in R^{N \times D} = R(I) \qquad (2)$$

$$Z_0 \in [x_{cls}; X_{patch}E] + E_{pos} \qquad (3)$$

### 3.2.2. Window-based Multi-head Self-attention with Shifted Windows

Within each block, MHA computes attention across multiple subspaces, consequently improving representational capacity relative to single-head attention. For an input token matrix (Z), query, key, and value projections are defined in equation 4 as

$$Q = ZW_Q, \quad K = ZW_K, \quad V = ZW_V, \qquad (4)$$

Scaled dot-product self-attention is processed in equation 5 as

$$SA(Q, K, V) = SoftMax\left(\frac{QK^T}{\sqrt{d_k}}\right) \cdot V \qquad (5)$$

where $d_k$ denotes the key dimensionality. Window-based attention is applied within local windows, and shifted windows are alternated across successive blocks, which enables cross-window interactions without full global quadratic cost, as described in the Swin formulation in equation 6. Residual learning is retained through the attention sub-layer:

$$Z' = Z + MHA(LN(Z)) \qquad (6)$$

### 3.2.3. Inverse Residual Block

The standard transformer feedforward network is replaced by a novel IRB to strengthen local pattern extraction and stabilize gradient flow. The conventional FFN in transformer models is commonly expressed using two linear layers through GELU activation functions:

$$FFN(X) = GELU(XW_1 + b_1)W_2 + b_2 \qquad (7)$$

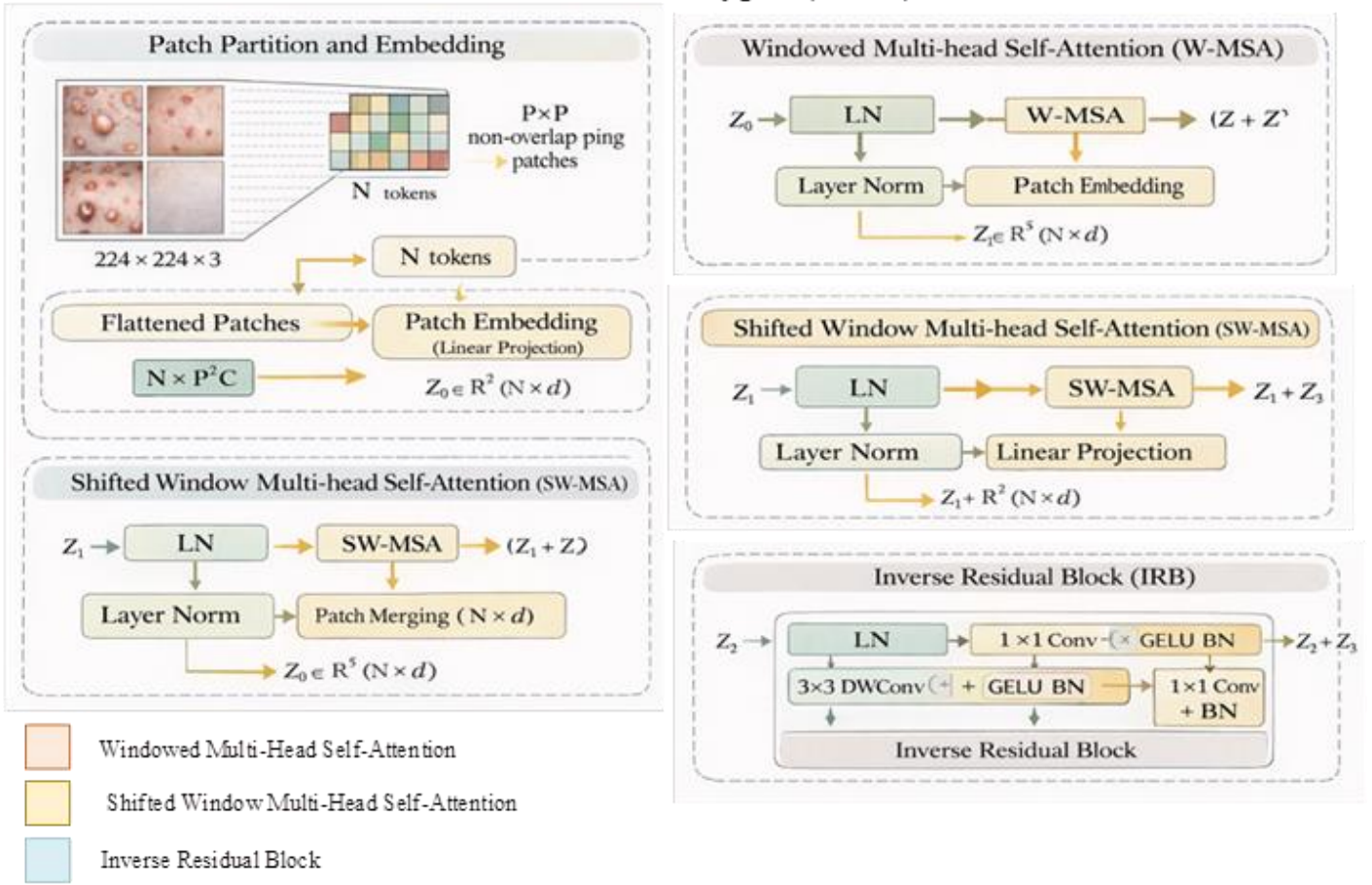

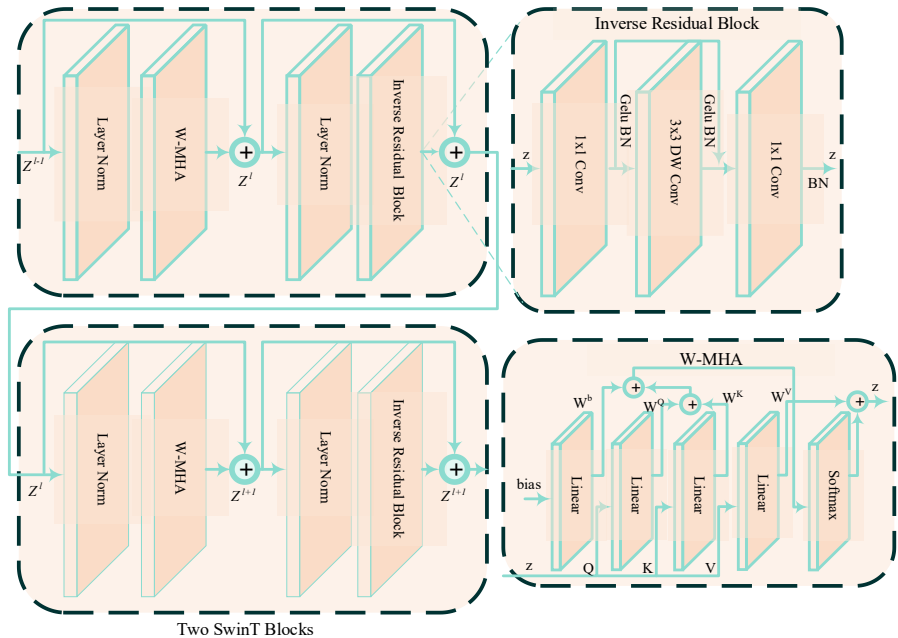

Figure 2. Customized Residual Swin Transformer (RSwinV2) Block.

The IRB adopts an inverted bottleneck structure with depth-wise convolution for efficient spatial aggregation. Expansion and projection operations are implemented through pointwise transforms around a depth-wise convolution, and a skip connection is retained to support optimisation stability in deep networks in equation 8. A compact formulation is:

$$\text{IRB}(\mathbf{X}) = \mathbf{X} + \prod(\text{DWConv}(\Phi(\mathbf{X}))) \qquad (8)$$

Where in equation 8 $\Phi(.)$ denotes channel expansion and $\prod(.)$ denotes projection back to the original dimensionality. The block output is formed with layer normalisation and a residual path:

$$\mathbf{Z}'' = \mathbf{Z}' + IRB\ (LN)(\mathbf{Z}') \qquad (9)$$

This design enables the simultaneous modeling of global dependencies via a shifted-window MHA and local lesion morphology via the convolutional IRB pathway, thereby supporting multi-class discrimination under visually similar rash conditions, as depicted in Equation 9.

## 4. Experimental Setup:

### 4.1. Dataset Details

This study utilizes a publicly accessible Kaggle dataset of dermatological images, professionally annotated for multi-class classification [36]. The set of images is divided into five groups, which are MPox, Measles, Cowpox, Cpox, and Normal. The sample images of each class are shown in Figure 3. The proportion of the images represented in each class, described in Table 2, simulates a real-world setup, which also shows class imbalance. This diversity and scale support the robust training and validation of DL models for differential diagnosis of MPox from visually similar conditions.

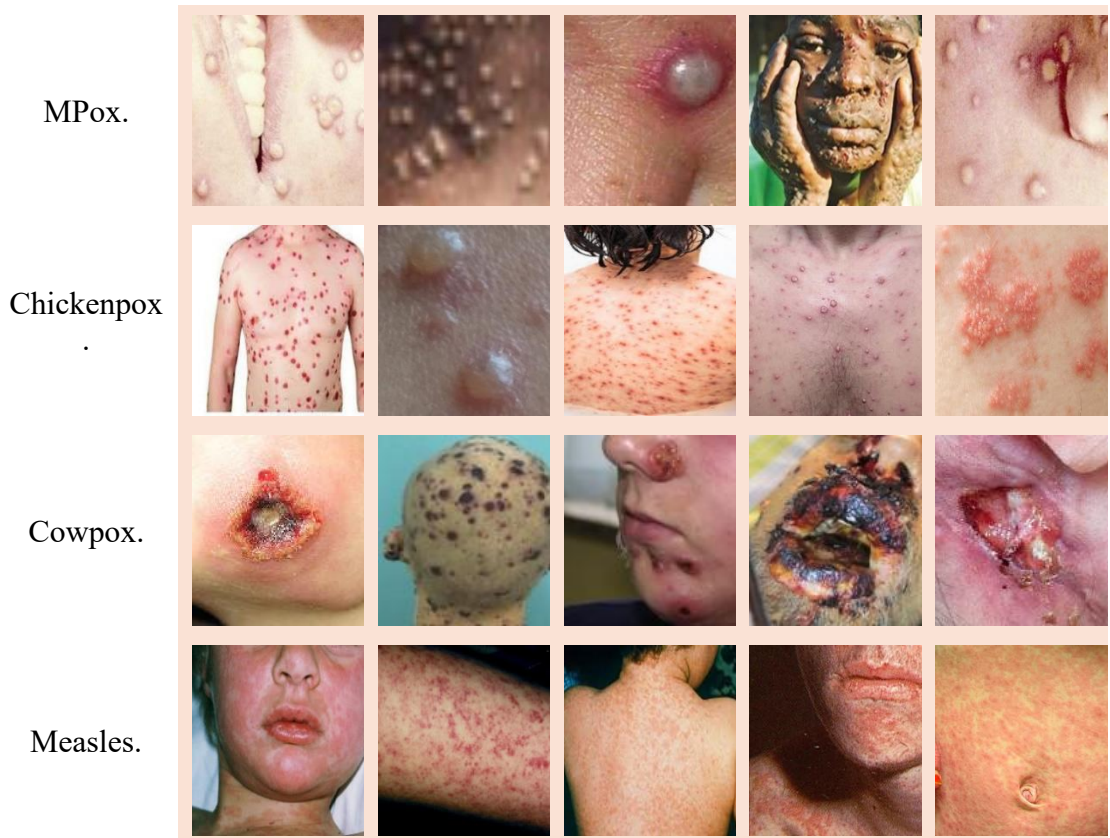

Normal. 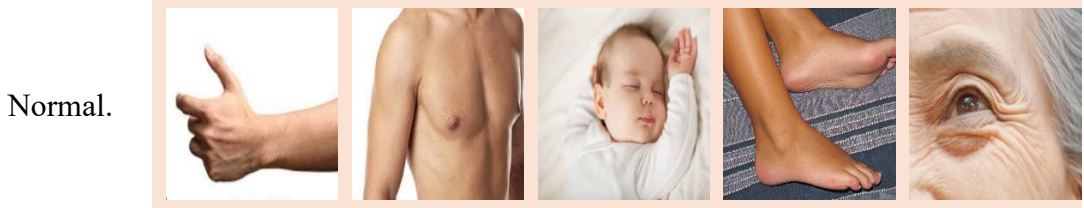

Figure 3. Five diagnostic categories: MPox, Measles, Cowpox, Chickenpox, and Normal skin Images.

Table 2. Composition of the benchmark dermatology image dataset.

| Characteristic | Specification |
|---|---|
| Total | 16630 |
| Class Distribution | |
| Measles | 1540 |
| Chickenpox | 2100 |
| Cowpox | 1850 |
| MPox | 7950 |
| Normal | 3190 |
| Train & Validation (80%) | (10643, 2,661) |
| Test set (20%) | (3,326) |
| Input Image | 224x 224 x 3 |

## 4.2. Experimental Setup

Training configuration for RSwinV2 employed Adam optimizer with a starting learning rate of $10^{-3}$, a weight decay parameter of 0.04, and a scheduled decay factor of 0.85 applied at 20-epoch intervals. The cross-entropy loss function was selected to manage inter-class imbalance. A batch size of 16 and a final-layer dropout of 0.3 were implemented to reduce overfitting risk. All experiments were coded in Python with TensorFlow and executed on hardware featuring an Intel Core i9-12$^{th}$ Gen CPU, 64 GB of RAM, and a NVIDIA GeForce RTX 4070 Ti GPU.

## 4.3. Evaluation Protocol

A hold-out validation protocol allocated 20% of the total data as a fixed test set. Model efficacy was quantified using conventional diagnostic metrics: Accuracy, Precision, Sensitivity, the F1-score, and the area under the receiver operating characteristic (AUC-ROC) and precision-recall curves (AUC-PR). These metrics are defined by conventional formulations, equations 10-13. Since there was paramount interest in the detection of the MPox cases, there was significant focus on maximizing the Sensitivity or Recall values. The standard error of Sensitivity was also determined, hence allowing the calculation of the 95% Confidence Interval using the z-test of 1.96.

$$Acc = \frac{TP+TN}{Total} \times 100 \qquad (10)$$

$$Sen = \frac{TP}{TP+FN} \times 100 \qquad (11)$$

$$Pre = \frac{TP}{TP+FP} \times 100 \qquad (12)$$

$$F - score = 2 \times \frac{Pre + Sen}{Pre + Sen} \qquad (13)$$

## 5. Results and Discussion

This section presents an analysis of the experimental outcomes and compares the efficiency of the proposed RSwinV2 network structure. Results will be compared to the advanced CNN, Vision Transformer, and hybrid CNN/Vision Transformer-based networks on both Kaggle and diverse datasets. Table 3 and Figures 4-5 illustrate the main assessment metrics, which involve Accuracy, Precision, Sensitivity, F1_Score, and AUC. Multi-class classification is carried out on the five classifications: MPox, Chickenpox, Cowpox, Measles, and Normal. The new model RSwinV2 produces the maximum accuracy of 96.51%, outperforming the CNN accuracy (94.77%) and the Hybrid LeViT model (96.09%). The ROC-AUC and PR-AUC scores are 0.9575 for ranking the classes accurately. The confusion matrix in Figure 4 demonstrates that the error differences are mostly around Cowpox and Chickenpox, stating that the visual similarity between the two classes is the major point of confusion here. Analysis on a per-class basis reveals that 96.51% of cases are properly classified for MPox, implying a misclassification error of 3.5%. The computational efficiency can be further noted based on comparisons regarding training complexity. The results revealed that RSwinV2 has lower training complexity compared to the employed CNN, ViT, and SwinT models, as presented in Table 3. The proposed approach has faster convergence compared to other alternatives, while hybrid models of CNN and ViT present an oscillatory nature during the optimization process.

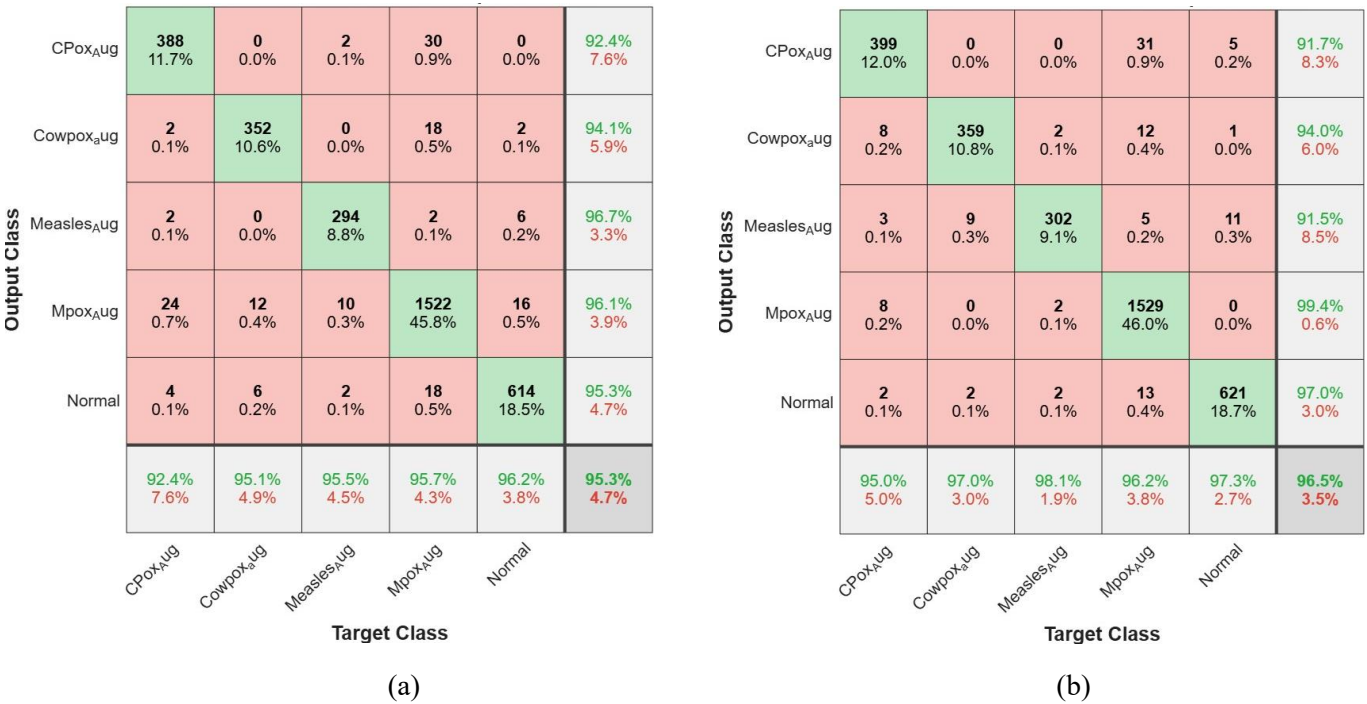

Figure 4. Confusion matrix for the evaluated methods: (a) the SwinT, and (b) the proposed RSwinV2 model.

Table 3. Model performance metrics for the proposed implementation configuration.

| Model | Acc % | Sen | Pre | F1 score |
|---|---|---|---|---|
| Efficient_Net_B0. | 85.81 | 0.8242 | 0.8678 | 0.8455 |
| Inception_ResNet_V2. | 86.17 | 0.8802 | 0.8214 | 0.8498 |
| Google_Net. | 90.08 | 0.8990 | 0.8745 | 0.8866 |
| Inception_V3. | 91.46 | 0.9095 | 0.9018 | 0.9056 |
| Mobile_Net_V4. | 92.18 | 0.9117 | 0.9167 | 0.9142 |
| DarkNet_53. | 93.21 | 0.9290 | 0.9161 | 0.9225 |
| VGG_16. | 94.41 | 0.9407 | 0.9317 | 0.9362 |
| ResNet_18. | 94.77 | 0.9406 | 0.9428 | 0.9417 |
| Cross_ViT. | 92.96 | 0.9137 | 0.9242 | 0.9189 |
| Local_ViT. | 93.21 | 0.9290 | 0.9161 | 0.9225 |
| Tiny_ViT. | 94.41 | 0.9407 | 0.9317 | 0.9362 |
| Mobile_ViT. | 95.13 | 0.9419 | 0.9434 | 0.9427 |
| Swin_T. | 95.31 | 0.9499 | 0.9493 | 0.9496 |
| **SwinV2**. | 96.03 | 0.9502 | 0.9582 | 0.9542 |
| **Proposed_RSwinV2** | **96.51** | **0.9621** | **0.9604** | **0.9613** |

Table 4. Performance comparison with prior research

| Models | Accuracy | Sensitivity | F1 Score |
|---|---|---|---|
| Existing_CNN's | | | |
| MobileNetv2 [32] | 91.11 | 0.9000 | 0.9000 |
| ResNet50 [31] | 82.96 | 0.8300 | 0.8400 |
| ShuffleNetV2 [37] | 79.00 | 0.5800 | 0.6700 |
| DarkNet-53 [20] | 85.78 | 0.8246 | 0.8420 |
| DenseNet201 [33] | 95.18 | 0.8982 | 0.8961 |
| MobileNet-V2 [15] | 95.40 | 0.9420 | 0.9400 |
| Existing_ViT's | | | |
| ViT's [22] | 93.00 | 0.9100 | 0.9200 |
| ViTB-18 [23] | 71.55 | 0.7926 | 0.6111 |
| ViT's [24] | 94.69 | 0.9500 | 0.9500 |
| ViT's [25] | 93.00 | 0.9300 | 0.9300 |
| Hybrid_Techniques | | | |
| Xception_CBAM_Dense [38] | 83.90 | 0.8910 | 0.9010 |
| VGG-16_(naïve_Bayes) [39] | 91.11 | - | - |
| 13 DL models and Ensemble method [40] | 87.13 | 0.8547 | 0.8540 |
| RestNet50 with TL [22] | 91.00 | 0.9000 | 0.9000 |
| Google-Net and Metaheuristic_Optimization [41] | 94.35 | 0.9500 | 0.9200 |
| Ensemble ViT's with Densenet-201 [26] | 81.91 | 0.7414 | 0.7816 |

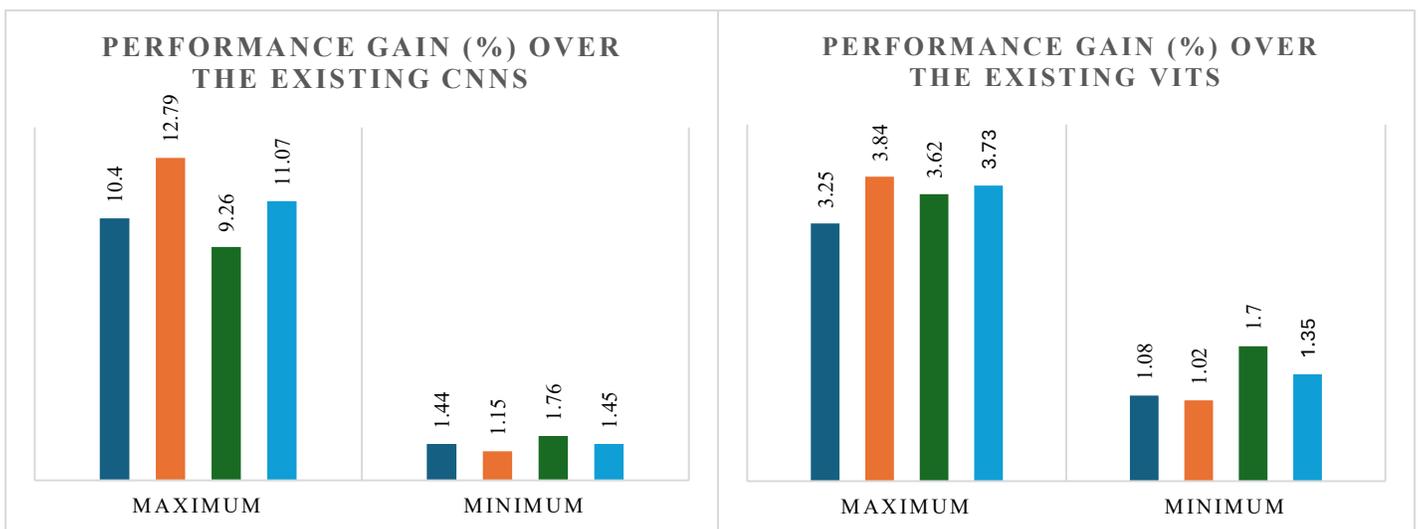

Figure 5. Performance improvements of the proposed RSwinv2 compared to established CNN and ViT.

## 5.1. PR/ROC Curve

PR-ROC curves are utilized to evaluate the discriminative ability of RSwinV2 on various thresholds. The PR and ROC curves on the five-class classification problem (MPox, chickenpox, cowpox, measles, and normal) are shown in Figure 6. They are plots that compare the predicted probabilities with the actual labels. The results are summarized and presented in Table 3 under the "20%" label setting. The PR-AUC and ROC-AUC scores for RSwinV2 are 0.9575, demonstrating high separability between classes. Compared to other baseline models, the relative improvements in PR-AUC and ROC-AUC range between 0.41% and 8.75%. ROC and PR curves are most preferable in the case of RSwinV2 compared to other models like CNN, ViT, and SwinT. The ROC points lie in the top-left portion, and the PR points lie in the top-right portion.

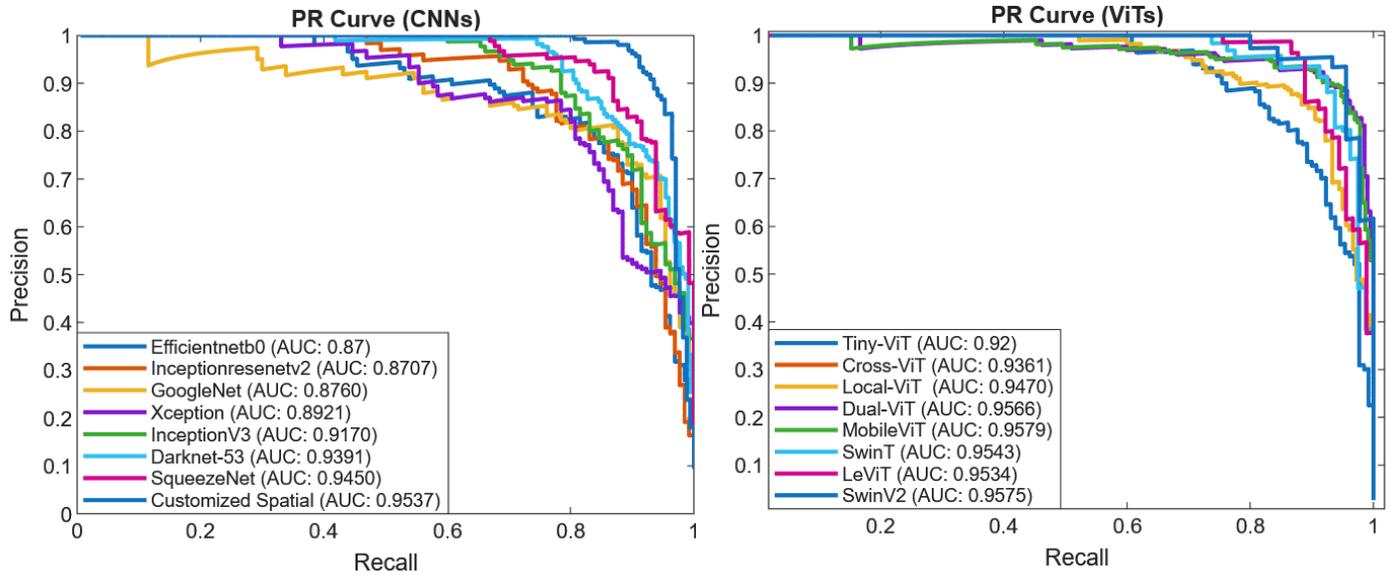

Figure 6. Detection rate analysis of proposed (RSwinV2) and existing CNNs and ViTs.

**5.2. Feature Space Visualization**

The feature representations will be investigated to enable the interpretation of the learned decision space. The penultimate feature set is obtained from the RSwinV2 model via a set-aside test set, and PCA is used to transform the feature embeddings of high dimensionality to two-dimensional feature spaces. The PCA projection on the two components for (a) the proposed RSwinV2, as well as (b) the LeViT baseline, has been represented in Figure 7. Greater separation among classes in the case of RSwinV2, concerning the MPox, chickenpox, cowpox, measles, and normal classes, as opposed to the overlap in the existing methodology in the PC1-PC2 direction, has been identified. Additionally, the separation in the first two components of the PCA projection supports the classification accuracy.

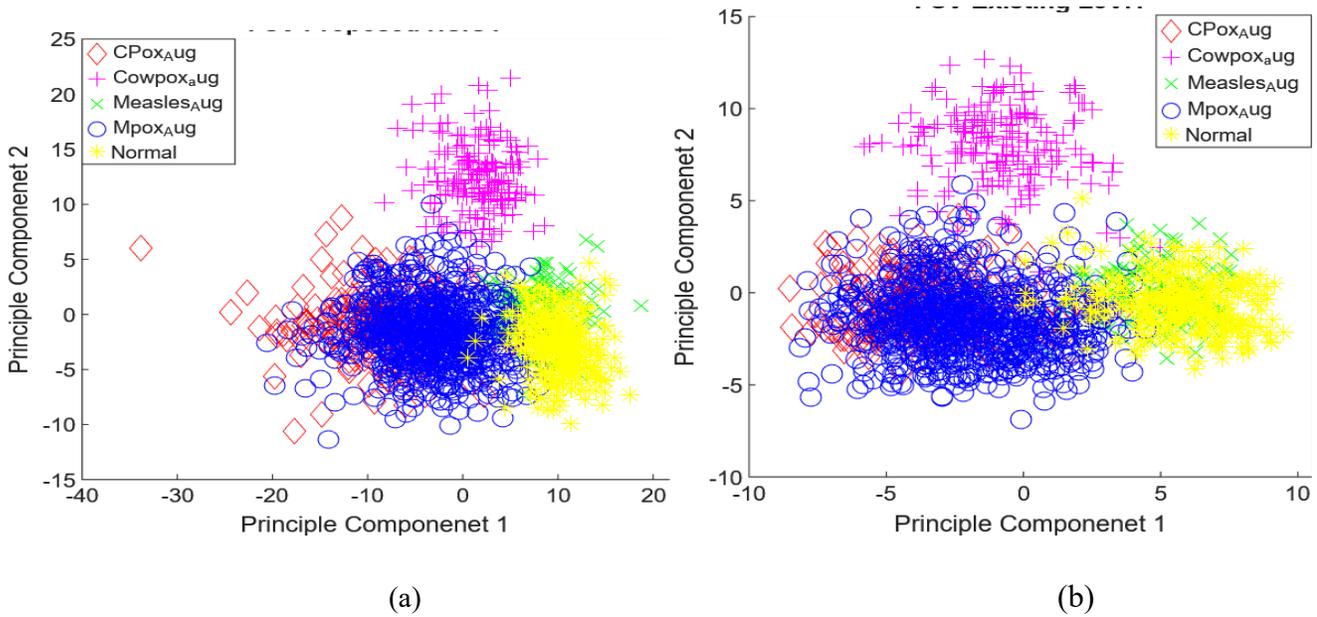

Figure 1. The PCA-based feature embeddings for: (a) the proposed RSwinV2, and (b) the LeViT baseline.

## 6. Conclusion

This work introduced RSwinV2, a tailored Swin Transformer framework designed to address key challenges in MPox image classification. The framework is designed with adaptations of a customized SwinTransformer framework regarding data dimension, embeddings, and a target-oriented result task applied for MPox image interpretation. It integrates patch and positional embeddings along with a multi-head attention method, enabling global dependency modeling with a newly developed Inverse Residual Block. The proposed model applies a hierarchical transformer-based architecture, dividing images into non-overlapping spatial patches. A sliding window-based self-attention method sustains interwindow connectivity while maintaining computational efficiency, thus reducing the strict locality requiremefnt of traditional Vision Transformers. The IRB applies convolution-based skip connections aiming to boost the identification of local correlated patterns of lesions and guarantee stable gradient flow. The proposed architecture of RSwinV2 helpfully sustains a balanced integration of global contextual cues relatedness with fine-grained local patterns, which significantly suppresses intra-class heterogeneities of MPox while improving inter-class discrimination boundaries against similar skin eruptive diseases like chickenpox, measles, and cowpox. Performance evaluation using a public Kaggle challenge indicates the effectiveness of the developed framework, offering RSwinV2 a high classification accuracy of 96.51% with a high F1 score of 96.13%. The RSwinV2 performs significantly better than conventional CNN models as well as SwinTransformer models. Future studies will aim towards the customization of such architectures to biomedical imaging areas involving challenging environments of low-contrast patterns, high heterogeneity of morphology, and limited availability of labeled training paradigms.

**Acknowledgment:**

We thank the Artificial Intelligence Lab, Department of Computer Systems Engineering, University of Engineering and Applied Sciences, for providing a healthy research environment and necessary computational resources.

# References

[1] O. Mitjà et al., "Monkeypox," *Lancet*, vol. 401, no. 10370, pp. 60–74, Jan. 2023, doi: 10.1016/S0140-6736(22)02075-X/ASSET/8EC43666-5BD1-46D5-9306-380DC7A858F3/MAIN.ASSETS/GR2.JPG.

[2] S. H. Khan, R. Iqbal, and S. Naz, "A Recent Survey of the Advancements in Deep Learning Techniques for Monkeypox Disease Detection," Nov. 2023, Accessed: Jan. 15, 2024. [Online]. Available: https://arxiv.org/abs/2311.10754v2

[3] H. Adler et al., "Clinical features and management of human monkeypox: a retrospective observational study in the UK," *Lancet Infect. Dis.*, vol. 22, no. 8, pp. 1153–1162, Aug. 2022, doi: 10.1016/S1473-3099(22)00228-6.

[4] S. H. Khan and R. Iqbal, "RS-FME-SwinT: A Novel Feature Map Enhancement Framework Integrating Customized SwinT with Residual and Spatial CNN for Monkeypox Diagnosis," Oct. 2024, Accessed: Apr. 30, 2025. [Online]. Available: https://arxiv.org/pdf/2410.01216

[5] A. Khan, S. H. Khan, M. Saif, A. Batool, A. Sohail, and M. Waleed Khan, "A Survey of Deep Learning Techniques for the Analysis of COVID-19 and their usability for Detecting Omicron," *J. Exp. Theor. Artif. Intell.*, pp. 1–43, Jan. 2023, doi: 10.1080/0952813X.2023.2165724.

[6] R. Iqbal and S. H. Khan, "RS-CA-HSICT: A Residual and Spatial Channel Augmented CNN Transformer Framework for Monkeypox Detection," Nov. 2025, Accessed: Jan. 03, 2026. [Online]. Available: https://arxiv.org/pdf/2511.15476

[7] B. Hill, "The 2022 multinational monkeypox outbreak in non-endemic countries," *https://doi.org/10.12968/bjon.2022.31.12.664*, vol. 31, no. 12, pp. 664–665, Jun. 2022, doi: 10.12968/BJON.2022.31.12.664.

[8] S. H. Khan et al., "COVID-19 detection and analysis from lung CT images using novel channel boosted CNNs," *Expert Syst. Appl.*, vol. 229, p. 120477, Nov. 2023, doi: 10.1016/j.eswa.2023.120477.

[9] A. Ullah et al., "Toward sustainable smart cities: applications, challenges, and future directions," *Int. J. Data Sci. Anal. 2025 207*, vol. 20, no. 7, pp. 6227–6247, Jun. 2025, doi: 10.1007/S41060-025-00856-2.

[10] S. H. Khan, "Malaria Parasitic Detection using a New Deep Boosted and Ensemble Learning Framework," *Converg. Inf. Ind. Telecommun. Broadcast. data Process. 1981-1996*, vol. 26, no. 1, pp. 125–150, Dec. 2022, doi: 10.1088/1674-1137/acb7ce.

[11] S. H. Khan, A. Sohail, A. Khan, and Y.-S. Lee, "COVID-19 Detection in Chest X-ray Images Using a New Channel Boosted CNN," *Diagnostics*, vol. 12, no. 2, p. 267, Jan. 2022, doi: 10.3390/diagnostics12020267.

[12] S. H. Khan, A. Sohail, M. M. Zafar, and A. Khan, "Coronavirus disease analysis using chest X-ray images and a novel deep convolutional neural network," *Photodiagnosis Photodyn. Ther.*, vol. 35, p. 102473, Sep. 2021, doi: 10.1016/j.pdpdt.2021.102473.

[13] M. M. Zahoor and S. H. Khan, "CE-RS-SBCIT A Novel Channel Enhanced Hybrid CNN Transformer with Residual, Spatial, and Boundary-Aware Learning for Brain Tumor MRI Analysis," Aug. 2025, Accessed: Jan. 03, 2026. [Online]. Available: https://arxiv.org/pdf/2508.17128

[14] W. Ullah, Y. N. Khalid, and S. H. Khan, "A Novel Deep Hybrid Framework with Ensemble-Based Feature Optimization for Robust Real-Time Human Activity Recognition," Aug. 2025, Accessed: Jan. 03, 2026. [Online]. Available: https://arxiv.org/pdf/2508.18695

[15] M. C. Irmak, T. Aydin, and M. Yaganoglu, "Monkeypox Skin Lesion Detection with MobileNetV2 and VGGNet Models," in *2022 Medical Technologies Congress (TIPTEKNO)*, IEEE, Oct. 2022, pp. 1–4. doi: 10.1109/TIPTEKNO56568.2022.9960194.

[16] D. Uzun Ozsahin, M. T. Mustapha, B. Uzun, B. Duwa, and I. Ozsahin, "Computer-Aided Detection and Classification of Monkeypox and Chickenpox Lesion in Human Subjects Using Deep Learning Framework," *Diagnostics*, vol. 13, no. 2, p. 292, Jan. 2023, doi: 10.3390/diagnostics13020292.

[17] D. Bala *et al.*, "MonkeyNet: A robust deep convolutional neural network for monkeypox disease detection and classification," *Neural Networks*, vol. 161, no. fibreculture, pp. 757–775, Apr. 2023, doi: 10.1016/j.neunet.2023.02.022.

[18] A. D. Raha *et al.*, "Attention to Monkeypox: An Interpretable Monkeypox Detection Technique Using Attention Mechanism," *IEEE Access*, vol. 12, pp. 51942–51965, 2024, doi: 10.1109/ACCESS.2024.3385099.

[19] M. M. Ahsan *et al.*, "Enhancing Monkeypox diagnosis and explanation through modified transfer learning, vision transformers, and federated learning," *Informatics Med. Unlocked*, vol. 45, p. 101449, Jan. 2024, doi: 10.1016/J.IMU.2024.101449.

[20] D. Biswas and J. Tešić, "Binarydnet53: a lightweight binarized CNN for monkeypox virus image classification," *Signal, Image Video Process.*, vol. 18, no. 10, pp. 7107–7118, Sep. 2024, doi: 10.1007/S11760-024-03379-8/METRICS.

[21] S. Naz and S. H. Khan, "Residual-SwinCA-Net: A Channel-Aware Integrated Residual CNN-Swin Transformer for Malignant Lesion Segmentation in BUSI," Dec. 2025, Accessed: Jan. 03, 2026. [Online]. Available: https://arxiv.org/pdf/2512.08243

[22] D. Kundu, U. R. Siddiqi, and M. M. Rahman, "Vision Transformer based Deep Learning Model for Monkeypox Detection," in *2022 25th International Conference on Computer and Information Technology (ICCIT)*, IEEE, Dec. 2022, pp. 1021–1026. doi: 10.1109/ICCIT57492.2022.10054797.

[23] T. Nayak *et al.*, "Deep learning based detection of monkeypox virus using skin lesion images," *Med. Nov. Technol. Devices*, vol. 18, p. 100243, Jun. 2023, doi: 10.1016/j.medntd.2023.100243.

[24] M. Aloraini, "An effective human monkeypox classification using vision transformer," *Int. J. Imaging Syst. Technol.*, Jul. 2023, doi: 10.1002/ima.22944.

[25] M. A. Arshed, H. A. Rehman, S. Ahmed, C. Dewi, and H. J. Christanto, "A 16 × 16 Patch-Based Deep Learning Model for the Early Prognosis of Monkeypox from Skin Color Images," *Comput. 2024, Vol. 12, Page 33*, vol. 12, no. 2, p. 33, Feb. 2024, doi: 10.3390/COMPUTATION12020033.

[26] G. Yolcu Oztel, "Vision transformer and CNN-based skin lesion analysis: classification of monkeypox," *Multimed. Tools Appl.*, vol. 83, no. 28, pp. 71909–71923, Aug. 2024, doi: 10.1007/S11042-024-19757-W/TABLES/3.

[27] A. Dosovitskiy *et al.*, "An Image is Worth 16x16 Words: Transformers for Image Recognition at Scale," *ICLR 2021 - 9th Int. Conf. Learn. Represent.*, Oct. 2020, Accessed: Oct. 13, 2023. [Online]. Available: https://arxiv.org/abs/2010.11929v2

[28] A. Khan *et al.*, "A Recent Survey of Vision Transformers for Medical Image Segmentation," Dec. 2023, Accessed: Jan. 15, 2024. [Online]. Available: https://arxiv.org/abs/2312.00634v2

[29] S. H. Khan and R. Iqbal, "A Comprehensive Survey on Architectural Advances in Deep CNNs: Challenges, Applications, and Emerging Research Directions," Mar. 2025, Accessed: Nov. 08, 2025. [Online]. Available: https://arxiv.org/pdf/2503.16546

[30] M. A. Arshad *et al.*, "Drone Navigation Using Region and Edge Exploitation-Based Deep CNN," *IEEE Access*, vol. 10, pp. 95441–95450, 2022, doi: 10.1109/ACCESS.2022.3204876.

[31] S. N. Ali *et al.*, "Monkeypox Skin Lesion Detection Using Deep Learning Models: A Feasibility Study," Jul. 2022, Accessed: Sep. 26, 2023. [Online]. Available: https://arxiv.org/abs/2207.03342v1

[32]  V. H. Sahin, I. Oztel, and G. Yolcu Oztel, "Human Monkeypox Classification from Skin Lesion Images with Deep Pre-trained Network using Mobile Application," *J. Med. Syst.*, vol. 46, no. 11, p. 79, Oct. 2022, doi: 10.1007/s10916-022-01863-7.

[33]  A. Sorayaie Azar, A. Naemi, S. Babaei Rikan, J. Bagherzadeh Mohasefi, H. Pirnejad, and U. K. Wiil, "Monkeypox detection using deep neural networks," *BMC Infect. Dis.*, vol. 23, no. 1, pp. 1–13, Dec. 2023, doi: 10.1186/S12879-023-08408-4/FIGURES/9.

[34]  E. H. I. Eliwa, A. M. El Koshiry, T. Abd El-Hafeez, and H. M. Farghaly, "Utilizing convolutional neural networks to classify monkeypox skin lesions," *Sci. Rep.*, vol. 13, no. 1, p. 14495, Sep. 2023, doi: 10.1038/s41598-023-41545-z.

[35]  G. M. Setegn and B. E. Dejene, "Explainable AI for Symptom-Based Detection of Monkeypox: a machine learning approach," *BMC Infect. Dis.*, vol. 25, no. 1, pp. 1–21, Dec. 2025, doi: 10.1186/S12879-025-10738-4/TABLES/9.

[36]  "Kaggle: Your Home for Data Science." Accessed: Oct. 13, 2023. [Online]. Available: https://www.kaggle.com/

[37]  T. Islam, M. A. Hussain, F. U. H. Chowdhury, and B. M. R. Islam, "Can Artificial Intelligence Detect Monkeypox from Digital Skin Images?," *bioRxiv*, p. 2022.08.08.503193, Oct. 2022, doi: 10.1101/2022.08.08.503193.

[38]  M. E. Haque, M. R. Ahmed, R. S. Nila, and S. Islam, "Human Monkeypox Disease Detection Using Deep Learning and Attention Mechanisms," in *2022 25th International Conference on Computer and Information Technology (ICCIT)*, IEEE, Dec. 2022, pp. 1069–1073. doi: 10.1109/ICCIT57492.2022.10055870.

[39]  V. Kumar, "Analysis of CNN features with multiple machine learning classifiers in diagnosis of monkeypox from digital skin images," *medRxiv*, p. 2022.09.11.22278797, Nov. 2022, doi: 10.1101/2022.09.11.22278797.

[40]  C. Sitaula and T. B. Shahi, "Monkeypox Virus Detection Using Pre-trained Deep Learning-based Approaches," *J. Med. Syst.*, vol. 46, no. 11, pp. 1–9, Nov. 2022, doi: 10.1007/S10916-022-01868-2/FIGURES/5.

[41]  A. H. Alharbi *et al.*, "Diagnosis of Monkeypox Disease Using Transfer Learning and Binary Advanced Dipper Throated Optimization Algorithm," *Biomimetics*, vol. 8, no. 3, p. 313, Jul. 2023, doi: 10.3390/biomimetics8030313.